\documentclass{article}

\usepackage{multicol}
\usepackage{multirow}
\usepackage[bookmarks=true]{hyperref}
\usepackage{booktabs}
\usepackage{amsmath,amsfonts,mathtools,amsthm}
\usepackage{upgreek}
\usepackage{tikz}
\usepackage{svg}
\usepackage{subcaption}

\usepackage{amssymb}  
\usepackage{pifont}
\usepackage{epstopdf}
\usepackage{float}
\newfloat{algorithm}{t}{lop}
\usepackage{stfloats}
\usepackage{mathtools}
\usepackage{algorithm}
\usepackage{algorithmicx}
\usepackage[noend]{algpseudocode}
\usepackage{siunitx}
\usepackage{verbatimbox}

\usepackage{enumitem}

\theoremstyle{remark}

\usepackage{calrsfs}
\DeclareMathAlphabet{\pazocal}{OMS}{zplm}{m}{n}
\DeclareMathAlphabet\mathbfcal{OMS}{zplm}{b}{n}

\usepackage{balance}

\newcommand{\figref}[1]{Fig.~\ref{#1}}

\usepackage{bm}

\usepackage{url}

\newcommand{\xxnote}[3]{}
\ifx\hidenotes\undefined
  \usepackage{color}
  \renewcommand{\xxnote}[3]{\color{#2}{#1: #3}}
\fi

\definecolor{green(munsell)}{rgb}{0.0, 0.66, 0.47}

\newcommand{\vg}{\mathbf{g}}
\newcommand{\vm}{\mathbf{m}}
\usepackage[preprint]{corl_2020} 

\title{Multimodal Trajectory Prediction via Topological Invariance for Navigation at Uncontrolled Intersections}

\author{
  Junha Roh$^{*}$, Christoforos Mavrogiannis$^{*}$, Rishabh Madan\thanks{Denotes equal contribution.} \\
  \textbf{Dieter Fox, Siddhartha S. Srinivasa} \\
  Paul G. Allen School, University of Washington, USA \\
  \texttt{\{rohjunha, cmavro, rishabhm, fox, siddh\}@cs.washington.edu} \\
}

\begin{document}
\maketitle


\begin{abstract}
We focus on decentralized navigation among multiple non-communicating rational agents at \emph{uncontrolled} intersections, i.e., street intersections without traffic signs or signals. Avoiding collisions in such domains relies on the ability of agents to predict each others' intentions reliably, and react quickly. Multiagent trajectory prediction is NP-hard whereas the sample complexity of existing data-driven approaches limits their applicability. Our key insight is that the geometric structure of the intersection and the incentive of agents to move efficiently and avoid collisions (rationality) reduces the space of likely behaviors, effectively relaxing the problem of trajectory prediction. In this paper, we collapse the space of multiagent trajectories at an intersection into a set of \emph{modes} representing different classes of multiagent behavior, formalized using a notion of topological invariance. Based on this formalism, we design \textit{Multiple Topologies Prediction} (MTP), a data-driven trajectory-prediction mechanism that reconstructs trajectory representations of high-likelihood modes in multiagent intersection scenes. We show that MTP outperforms a state-of-the-art multimodal trajectory prediction baseline (MFP) \citep{tang2019multiple} in terms of prediction accuracy by 78.24\% on a challenging simulated dataset. Finally, we show that MTP enables our optimization-based planner, MTPnav, to achieve collision-free and time-efficient navigation across a variety of challenging intersection scenarios on the CARLA simulator.  A link to our code implementation and demo videos can be found at  \href{https://sites.google.com/view/multiple-topologies-prediction}{this} website.

\end{abstract}




\section{Introduction}\label{sec:introduction}

The widespread interest in autonomous driving technology in recent years \citep{McKinsey} has motivated extensive research in multiagent navigation in driving domains. One of the most challenging driving domains \citep{fhwa} is the \emph{uncontrolled} intersection, i.e., a street intersection that features no traffic signs or signals. Within this domain, we focus on scenarios in which agents do not communicate explicitly or implicitly through e.g., turn signals.
This model setup gives rise to challenging multi-vehicle encounters that mimic real-world situations (arising due to human distraction, violation of traffic rules or special emergencies) that result in fatal accidents \citep{fhwa}.
The frequency and severity of such situations has motivated vivid research interest in uncontrolled intersections \citep{isele18,mavrogiannis2020implicit,dfk20}.

In the absence of explicit traffic signs, signals, rules or explicit communication among agents, avoiding collisions at intersections relies on the ability of agents to predict the dynamics of interaction amongst themselves. One prevalent way to model multiagent dynamics is via trajectory prediction. However, multistep multiagent trajectory prediction is NP-hard \citep{cooper90}, whereas the sample complexity of existing learning algorithms effectively prohibits the extraction of practical models. Our key insight is that the geometric structure of the intersection and the incentive of agents to move efficiently and avoid collisions with each other (rationality) compress the space of possible multiagent trajectories, effectively simplifying inference. 

\begin{figure}
\centering
\includegraphics[width=\linewidth]{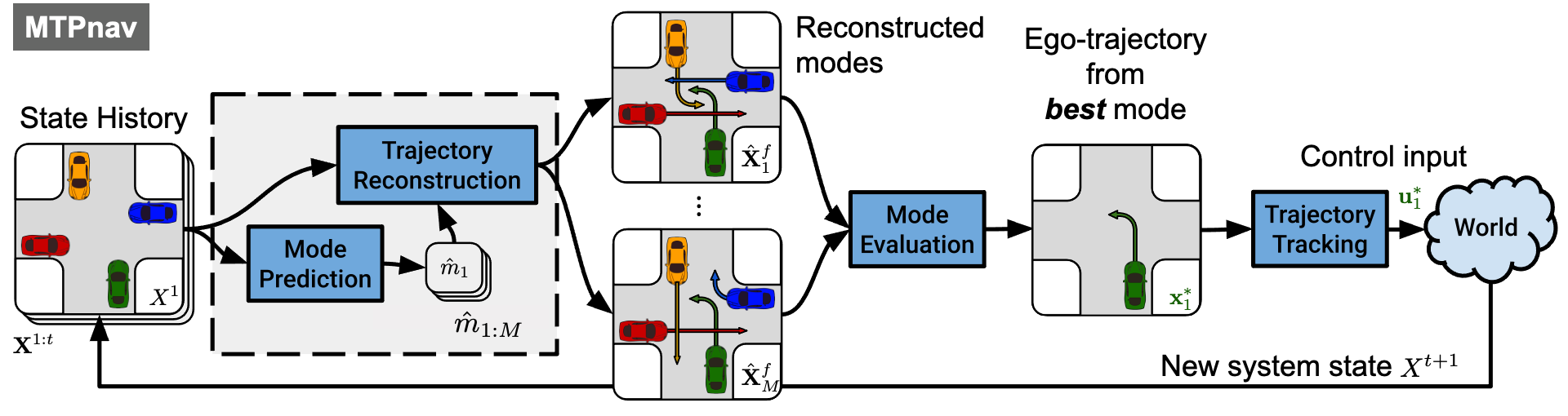}
\caption{\textbf{Overview of our decentralized navigation planner, MTPnav:} The ego-vehicle (shown in green) invokes MTP (Multiple Topologies Prediction), our trajectory prediction mechanism, which determines $M$ \emph{modes} of highly likely future multiagent behavior, and reconstructs trajectory representatives for them.
These modes are then evaluated based on the quality of their representatives incorporating considerations such as efficiency, clearance, likelihood, and smoothness. The ego-trajectory from the best mode is then passed to a controller which tracks the first waypoint.}
\label{fig:overview}
\end{figure}

In this paper, we explicitly account for this structure by collapsing the space of multiagent trajectories at an intersection into a finite set of \emph{modes} of joint behavior. We represent modes using a notion of topological invariance \citep{Berger2001invariants}: all trajectories belonging to the same mode leave the same topological signature (\figref{fig:execution-example2} depicts an example). Leveraging the flexibility and expressiveness of Graph Neural Networks \citep{battaglia2018relational}, we present a trajectory-prediction architecture that biases inference towards high-likelihood modes. We show that our architecture, entitled \emph{Multiple Topologies Prediction} (MTP), outperforms a state-of-the-art baseline (MFP \citep{tang2019multiple}) by 78.24\% in terms of reconstruction accuracy on a dataset of challenging intersection crossing tasks. Based on MTP, we design MTPnav, an optimization-based planning framework (see \figref{fig:overview}) which achieves time-efficient, collision-free navigation across a variety of challenging multiagent intersection-crossing scenarios in the CARLA simulation environment \citep{Dosovitskiy17}.

\section{Related Work}\label{sec:relatedwork}


In recent years, the end goal of autonomous driving has motivated extensive research on prediction and planning for navigation in multiagent domains. Considerable attention focuses on the task of multiagent trajectory prediction. Earlier works in the area employed physics-based models \citep{ammoun2009real}, hidden Markov models \citep{firl2012predictive}, and dynamic Bayesian networks \citep{gindele2015learning}. With the advent of deep learning and the availability of several public trajectory datasets \citep{Argoverse,lyft2020}, many  recent works are using Recurrent Neural Networks \citep{Gupta18,Deo18,tang2019multiple} and graph-based models \citep{Li19,chandra19,li20}. For instance, \citet{Deo18} use social pooling layers along with maneuver-based trajectory generation by labeling trajectories with maneuvers, which leads to multimodal predictions. \citet{Li19} make use of graph-convolutional networks combined with an LSTM-based encoder-decoder to account for inter-vehicle interactions. A recent line of work employs models for \emph{multimodal} trajectory prediction in an effort to account for the uncertainty in multiagent navigation \citep{SchmerlingLeungEtAl2018,MavKne_WAFR18,tang2019multiple,liang2019garden,monti2020dagnet}. For example, \citet{tang2019multiple} propose the Multiple Futures Prediction (MFP) method for multimodal multiagent trajectory predictions by capturing different behaviors present in the data using discrete latent variables in an unsupervised manner. Our work also embraces the virtues of multimodality but differs by  incorporating a mathematically sound, compact and interpretable formalism for representing multimodality using concepts from topology to drive the learning process.

In parallel, relevant works on planning and control are focusing on integrating models of multiagent dynamics in the decision making. Some works employ implicit models of interaction, such as risk level sets~\citep{Pierson_ICRA18}, deep reinforcement learning~\citep{isele18}, or imitation learning~\citep{rhinehart2018deep}. While such approaches may yield desirable performance in interesting driving scenes, they abstract away the richness of interaction that unfolds in driving domains, thus showing limited applicability, and they are often further constrained by data dependencies. In an effort to explicitly model interaction, some works model multiagent dynamics as \emph{games} by employing tools from game theory~\citep{Cleach-RSS-20,dfk20}. While game-theoretic approaches elegantly capture the richness of interactions in driving domains, the challenge of computing (Nash) equilibria and the intractability of scaling to large numbers of agents limits their applicability. Other works have employed topological abstractions such as braids~\citep{mavrogiannis2020implicit,artin} to enable agents to reason over multiagent collision-avoidance strategies. Topological braids capture critical events in multiagent navigation, such as the order of passing maneuvers, in a compact and interpretable fashion \citep{mavrogiannis_ijrr,mavblukne_iros2017}; however, their lack of analytic descriptions complicates the construction of generalizable prediction models and so limits their applicability. Instead of braids, our work leverages a notion of topological invariance \citep{Berger2001invariants} with analytic descriptions that enable the use of data-driven methods. This allows us to retain the benefits of topological reasoning while achieving state-of-the-art performance in challenging and realistic multiagent intersection-crossing tasks.
\section{Problem Statement}\label{sec:problemstatement}

Consider a four-way uncontrolled intersection (see \figref{fig:execution-example2}), where no explicit rules (e.g., right-turn priority) or signals (e.g., traffic lights) are set in place to regulate traffic. Assume that $1< n \leq 4$ non-communicating agents with \emph{simple-car} kinematics are navigating. Denote by $x_{i}\in\pazocal{X}\subseteq SE(2)$ the state of agent $i\in \pazocal{N} = \{1,\dots, n\}$ with respect to (wrt) a fixed reference frame. Agent $i$ is tracking a path $\tau_{i}:I\to \pazocal{X}$, for which it holds that $\tau_{i}(0) = s_{i}$ and $\tau_{i}(1) = d_{i}$, where $s_{i}, d_{i}\in \pazocal{X}$ are states lying at different sides of the intersection (i.e., right, left, up, down) and $I=[0,1]$ is a path parameterization. At timestep $t$ (assume a fixed time discretization $dt$), agent $i$ executes a policy $\pi_{i}:\pazocal{X}\to\pazocal{U}_{i}$ that generates controls $u_{i}\in \pazocal{U}_{i}$ (speed and steering angle) contributing progress along $\tau_{i}$ while avoiding collisions with $j\neq i$. Agent $i$ is not aware of the intended path $\tau_{j}$, the destination $d_{j}$, or the policy $\pi_{j}$ of agent $j\neq i\in \pazocal{N}$.



Let agent \#1 be the \emph{ego} agent. The ego agent perfectly observes the current system state $X^{t} = (x^{t}_{1},\dots, x^{t}_{n})\in\pazocal{X}^n$ and has access to the complete system state history $\mathbf{X}^{1:t} = (X^{1},\dots, X^{t})$. Our goal is to design a policy $\pi_{1}$ that enables the ego agent to follow collision-free and time-efficient intersection crossings despite the uncertainty resulting from the constraint of no communication.

\section{Navigation with Multiple Topologies Prediction}\label{sec:framework}


We describe a policy $\pi_{1}$ for decentralized navigation at uncontrolled intersections. Our policy is based on a data-driven mechanism for multimodal multiagent trajectory prediction. Our mechanism leverages a topological formalism that effectively compresses the space of possible multiagent trajectories into a set of \emph{modes} of multiagent behavior. This allows us to guide trajectory prediction towards high-likelihood modes. At planning time, our policy evaluates a set of high-likelihood multiagent trajectory alternatives and follows the ego trajectory from the one with minimum cost.






\subsection{A Topological Formalism of Modes for Navigation at Intersections}\label{sec:winding}

\begin{figure}
\centering
\begin{subfigure}{0.32\linewidth}
\centering
\includegraphics[width = \linewidth, trim = {.1cm .1cm .1cm .1cm},clip]{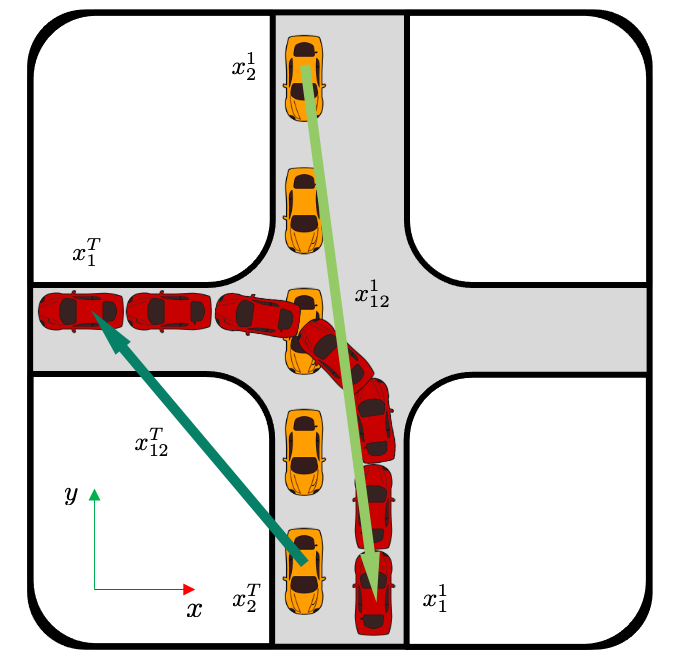}
\caption{Top view of agents' trajectories.\label{fig:execution-example2}}
\end{subfigure}
~
\begin{subfigure}{0.32\linewidth}
\centering
\includegraphics[width = \linewidth, trim = {.1cm .1cm .1cm .1cm},clip]{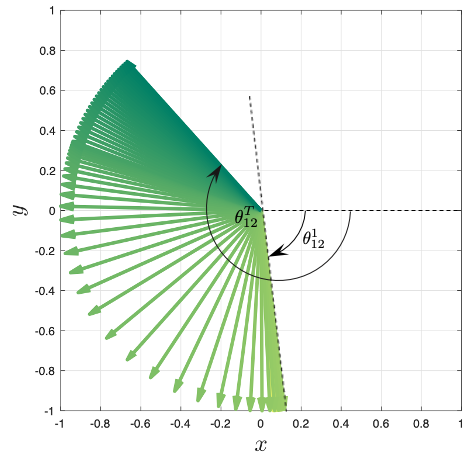}
\caption{Unit winding vector trajectory.\label{fig:windingvector2}}
\end{subfigure}
~
\begin{subfigure}{0.32\linewidth}
\centering
\includegraphics[width = \linewidth, trim = {.1cm .1cm .1cm .1cm},clip]{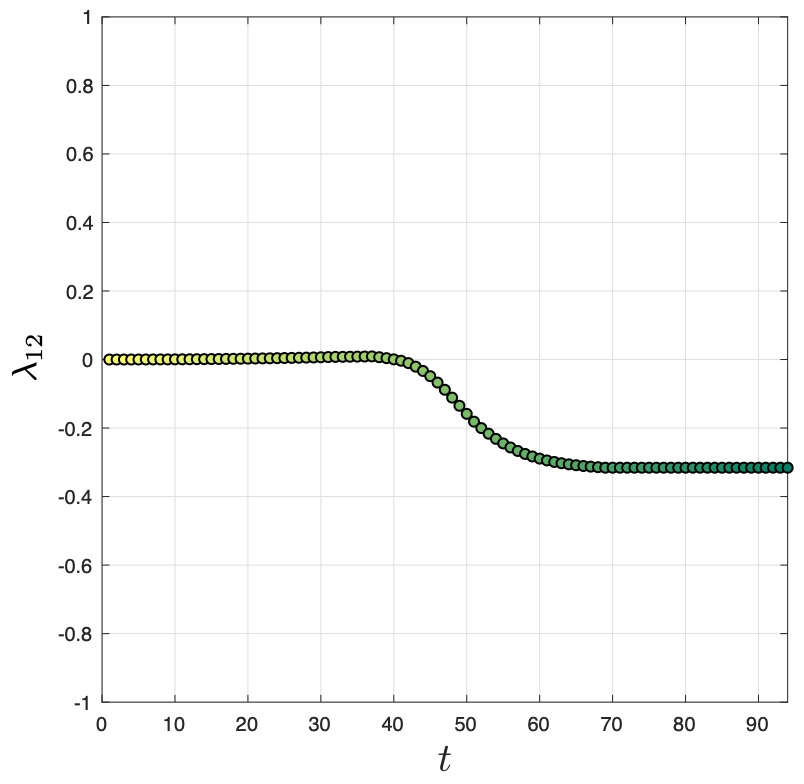}
\caption{Winding number convergence.\label{fig:windingnumber2}}
\end{subfigure}
\caption{\textbf{Identifying modes of intersection crossing:} (\subref{fig:execution-example2}) top view of agents' trajectories, indicating the initial and terminal winding vectors; (\subref{fig:windingvector2}) unit winding vector trajectory (darkness increases with time); and (\subref{fig:windingnumber2}) winding number convergence. Any execution of this scenario in which the orange agent passes before the red agent converges to a negative winding number. \label{fig:windingnumberexample}}
\end{figure}

Consider two agents navigating a four-way intersection. Denote by $x_{12}^{t} = x_{1}^{t}-x_{2}^{t}$ a vector--referred henceforth as the \emph{winding vector}-- expressing the relative location of agents and by $\theta_{12}^{t}$ its angle with respect to a global frame at timestep $t\geq 1$ (see \figref{fig:execution-example2}). From time $t$ to $t+1$, agents' displacement, $\Delta x^{t}_{12} = x_{12}^{t+1} - x_{12}^{t}$, results in a rotation $\Delta \theta^{t}_{12} = \theta_{12}^{t+1} - \theta_{12}^{t}$. Taking the sum of these rotations from the beginning of the execution ($t=1$) until the end ($t=T$), we derive a \emph{winding number} \citep{Berger2001invariants} characterizing the total relative rotation of both agents:
\begin{equation}
    \lambda_{12} = \frac{1}{2\pi} \sum_{t=1}^{T-1}\Delta\theta_{12}^{t}\mbox{.}
\end{equation}
The value $\lambda_{12}$ tracks the signed number of times agents $1,2$ ``revolved" around each other from time $t=1$ to $t=T$. 

The winding number is a \emph{topological invariant}: fixing agents' endpoints $s_{1}$, $s_{2}$, $d_{1}$, $d_{2}$, any topology-preserving deformations of agents' trajectories (continuous trajectory deformations not involving penetrations or cuts) would be identified using the same winding-number value $\lambda_{12}$. Importantly, the sign of the winding number $w_{12} = \mathrm{sgn}(\lambda_{12})$ indicates the side on which the two agents pass each other. A right-side passing corresponds to a clockwise rotation of the winding vector $x_{12}$, yielding a positive winding number $\lambda_{12}>0$; a left-side passing corresponds to a counterclockwise rotation of the winding vector $x_{12}$, yielding a negative winding number $\lambda_{12}<0$. \figref{fig:windingnumberexample} illustrates the machinery of the winding number at an intersection example with two agents. 

Extending this idea to a scene with $n$ agents with endpoints $S = (s_{1}, \dots, s_{n})$ and $D = (d_{1},\dots, d_{n})$, we define the \emph{topology} of an intersection-crossing task as:
\begin{equation}
    W = \left(\dots, w_{ij}, \dots\right), ij\in\pazocal{N}_{e}\mbox{,}
\end{equation}
where $\pazocal{N}_{e}$ is the set of all pairs formed among $n$ agents. Further, we define a \emph{mode} of intersection crossing as a tuple $m = (S, D, W)$. At an intersection-crossing task, a mode encodes \emph{where} agents are heading (destination $D$) and \emph{how} they are getting there (topology $W$) from their initial state, $S$. 



\subsection{Multiple Topologies Prediction}


Leveraging the formalism of modes, we build \emph{Multiple Topologies Prediction} (MTP), a multimodal trajectory prediction framework. MTP biases trajectory inferences towards a set of high-likelihood modes. Given a recent state history $\mathbf{X}^{p} = \mathbf{X}^{t_{p} + 1:t}$ of $h_{p} = t - t_{p}$ timesteps in the past, MTP: 1) determines a set of highly likely modes of future multiagent behavior $\hat{\vm} = \{\hat{m}_1, \cdots, \hat{m}_M\}$; 2) predicts a corresponding set of trajectory representatives $\hat{\mathbfcal{X}} = \{\hat{\mathbf{X}}_1^f, \dots, \hat{\mathbf{X}}_M^f\}$ where $\hat{\mathbf{X}}_l^f = \mathbf{X}^{t + 1:t_f}_{l}$, $l = 1, \dots, M$, and $h_{f} = t_{f} - t$ is a horizon into the future. We construct models for the two stages of prediction using a GraphNet architecture \citep{battaglia2018relational}. We first describe how to reconstruct a trajectory for a desired mode $m$ and then describe a technique for generating high-likelihood modes.


\subsubsection{Mode-Conditioned Trajectory Reconstruction}\label{sec:graphnet}

\begin{figure}
    \centering
    \includegraphics[width = \linewidth]{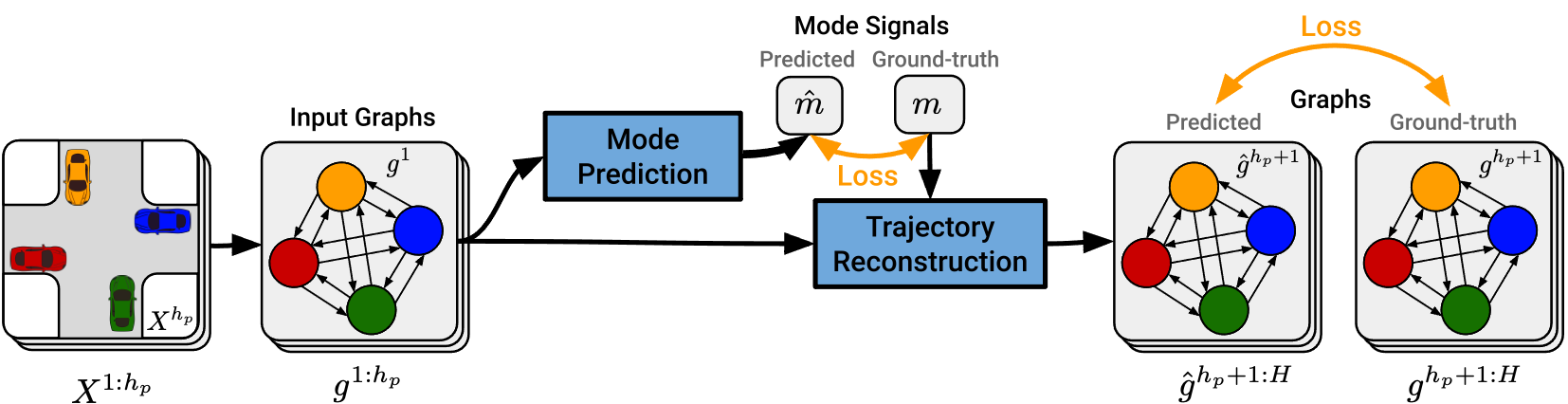}
    \caption{\textbf{Overview of model training.} We train a mode prediction and a trajectory reconstruction model, using the GraphNet \citep{battaglia2018relational} framework. Encoding the system state as a graph, we supervise predicted mode signals and reconstructed trajectories.}
    \label{fig:training}
\end{figure}

Our model represents the world state at time $t$, $X^t$ as a directed graph $g^t$ (see \figref{fig:training}).
The graph $g^t$ consists of a set of node attributes $V = \left\{ \mathbf{v}_i^t : i \in \pazocal{N} \right\}$, a set of edge attributes and corresponding node indices $E = \left\{ \left( \mathbf{e}_k^t, s_k, r_k \right) : k \in \pazocal{N}_e, s_k, r_k \in \pazocal{N} \right\}$, and a global attribute $\mathbf{u}^t$.
We set the position and velocity of agent $i$ at time $t$ to be a node attribute, i.e., $\mathbf{v}_i^t = \left( x_i^t, \Delta x_i^t \right)$, and the relative position of the $k$-th pair of agents at time $t$ to be an edge attribute, i.e., $\mathbf{e}_k^t = \Delta x_{s_k r_k}^t$.
We initialize the global attribute with a zero vector since the model does not have a dynamic global context. However, we leave the global attribute for the model to have ability to keep aggregated information in the attribute.

\begin{figure}
    \centering
    \begin{subfigure}{0.35\linewidth}
    \centering
    \includegraphics[width=\linewidth]{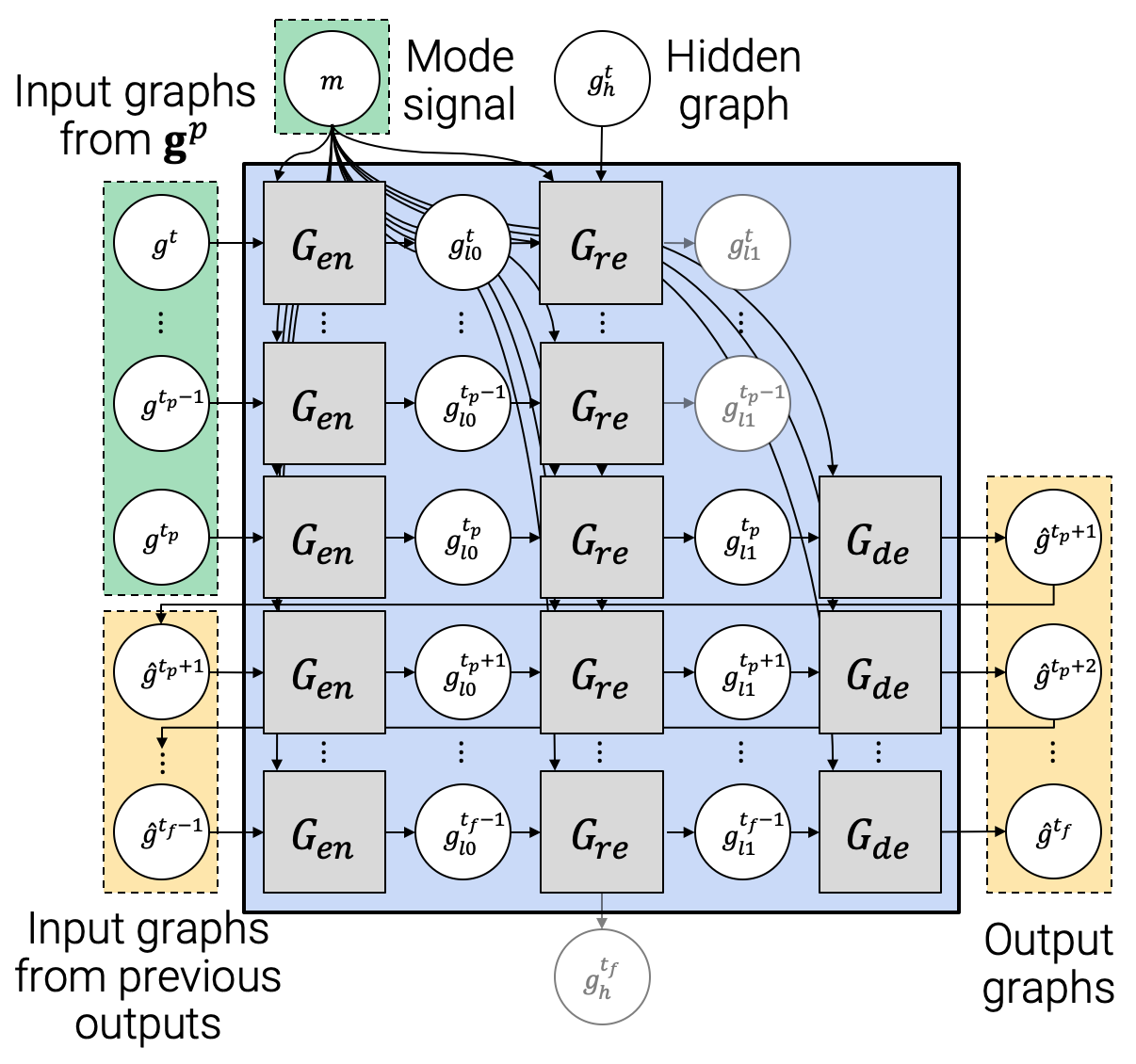}
    \caption{Trajectory reconstruction model.}
    \label{fig:traj-recon-model}
    \end{subfigure}
    ~
    \begin{subfigure}{0.58\linewidth}
    \centering
    \includegraphics[width=\linewidth]{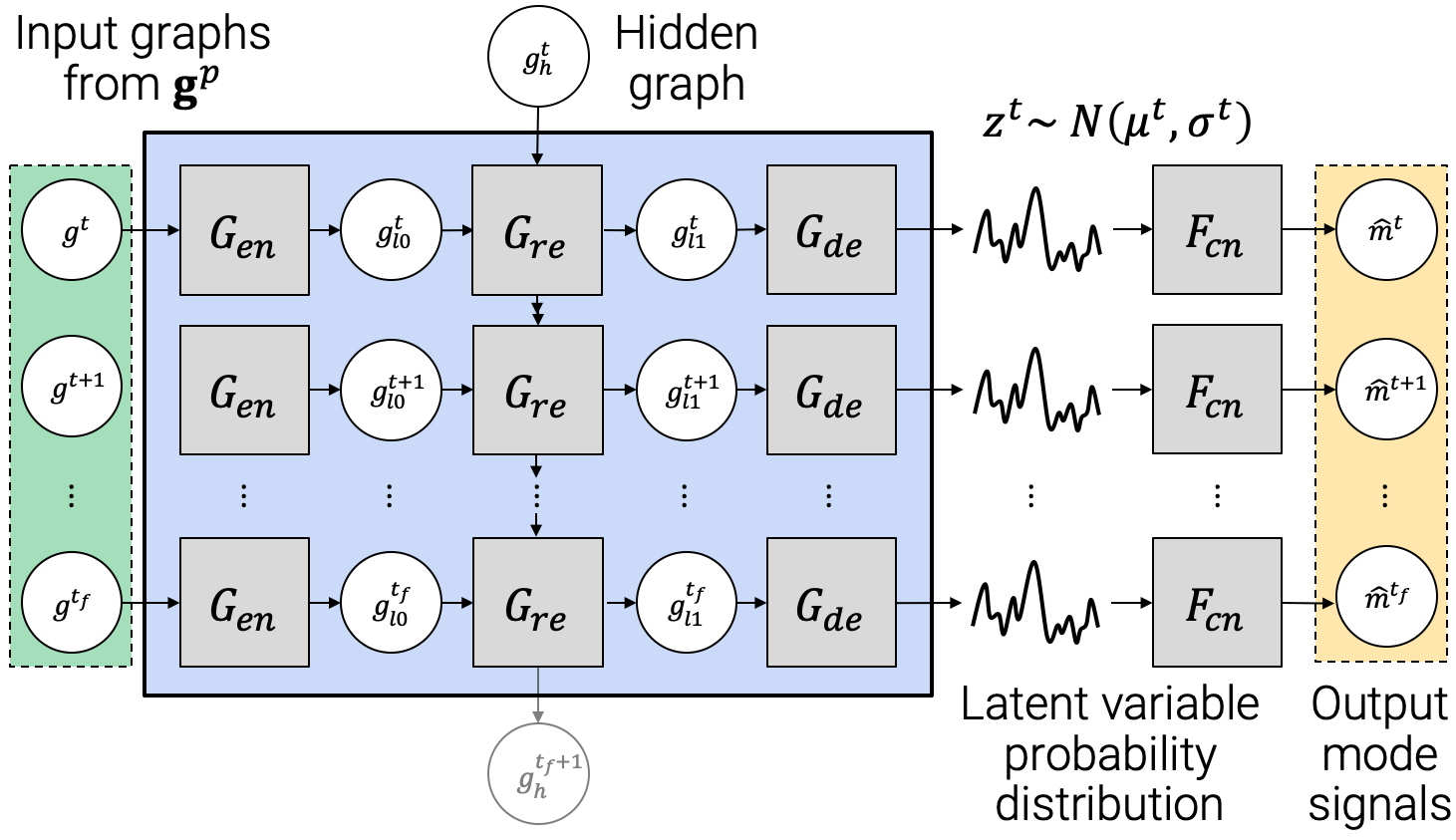}
    \caption{Mode prediction model.}
    \label{fig:mode-pred-model}
    \end{subfigure}
    
    \caption{\textbf{Internal topology of trajectory reconstruction (\subref{fig:traj-recon-model}) and mode prediction (\subref{fig:mode-pred-model}) models.} At time $t$, the trajectory reconstruction model takes as input the graph $g^t$, a hidden graph $g_h^t$, and a mode $m$ and generates an output graph $\hat{g}^{t+1}$. This procedure is repeated up to timestep $t_{f}$, yielding a trajectory in the form of a sequence of graphs $\hat{\vg}^f = \hat{g}^{t+1:t_{f}}$.
    The mode prediction model takes as input the graph $g^t$ and a hidden graph $g_h^t$ and produces a probability distribution over a latent variable $\mathbf{z}^t$.
    A sample is drawn from the distribution and transformed into a predicted mode $\hat{m}^t$.
    Variables in green shades indicate input from the ground-truth in training,  variables in yellow shades are the outputs, and grey variables are discarded.
    At inference, graphs up to current timestep $t$, $g^{t_p:t}$, are fed to the mode-prediction model to estimate modes, $\left\{ \hat{m}_{1:M} \right\}$. Given estimated modes and the graphs, the trajectory-reconstruction model predicts future trajectories.\label{fig:graph_net_compositions}
    }
\end{figure}

Our model takes as input the world state $g^t$ and outputs a predicted world state $\hat{g}^{t+1}$.
Computation takes place within a network of interconnected blocks: an encoder block $G_{\textrm{en}}$, a decoder block $G_{\textrm{de}}$, and a recurrent block $G_{\textrm{re}}$ (see \figref{fig:traj-recon-model}), introduced to handle temporal data effectively.
The block $G_{\textrm{en}}$ encodes the graph $g^t$ into a \emph{latent} graph $g_{l0}^{t}$.
Then, the recurrent block $G_{\textrm{re}}$ modifies it into $g^{t}_{l1}$ and passes it to $G_{\textrm{de}}$, which returns an updated graph $\hat{g}^{t+1}$ that represents the predicted state of the world in Cartesian coordinates. Within each block, the following computations take place (we set $g = g^{t}$, $g' = g^{t+1}$, and assume $i \in \pazocal{N}, k \in \pazocal{N}_e$ for brevity):
\begin{equation}
\label{eq:graph-update}
\begin{aligned}
    \mathbf{e}_k'           &= \mathbf{e}_k + \phi_e \left( \mathbf{e}_k, \mathbf{v}_{s_k}, \mathbf{v}_{r_k}, \mathbf{u}, \mathbf{w}_k \right) & 
    \bar{\mathbf{e}}_i'     &= \rho_{e \rightarrow v} \left( E_i' \right) &
    \quad E_i'              &= \left\{ \left( \mathbf{e}_k', s_k, r_k \right), r_k = i \right\} \\ 
    \mathbf{v}_i'           &= \mathbf{v}_i + \phi_v \left( \bar{\mathbf{e}}_i', \mathbf{v}_i, \mathbf{u}, \mathbf{d}_i \right) & 
    \bar{\mathbf{e}}'       &= \rho_{e \rightarrow u} \left( E' \right) &
    V'                      &= \left\{ \mathbf{v}_i' \right\} \\ 
    \mathbf{u}'             &= \mathbf{u} + \phi_u \left( \bar{\mathbf{e}}', \bar{\mathbf{v}}', \mathbf{u} \right) & 
    \bar{\mathbf{v}}'       &= \rho_{v \rightarrow u} \left( V' \right) &
    E'                      &= \left\{\left( \mathbf{e}_k', s_k, r_k \right)\right\},
\end{aligned}
\end{equation}

where $\phi$ and $\rho$ are respectively update and aggregation functions, $\mathbf{d}_i \in \left\{0, 1 \right\}^4,      \mathbf{w}_k \in \left\{ 0, 1 \right\}^2$ are one-hot encoded vectors of the destination of agent $i$ and of the sign of the winding number of edge $k$ respectively.
We model update functions $\phi_{e}$, $\phi_{v}$, $\phi_{u}$ using fully connected layers with ReLU and layer normalization and aggregation functions $\rho_{e\to v}$, $\rho_{e\to u}$, $\rho_{v \to u}$ as means over the updates of their corresponding sets.
In the recurrent block, $G_{\textrm{re}}$, we use a single-layer GRU (Gated Recurrent Unit) \citep{cho-etal-2014-properties} to efficiently infer the sequence of vehicle positions.

    

\subsubsection{Sampling High-Likelihood Modes}\label{sec:sampling}

The trajectory reconstruction model takes a mode $m$ as input. The number of modes is exponential in the number of agents, and not all modes are equally likely.  
To address this issue, we sample high-likelihood modes by adapting our reconstruction architecture (see \figref{fig:traj-recon-model}) into a mode prediction model (see \figref{fig:mode-pred-model}).
The model fits a Gaussian distribution over a latent variable $\mathbf{z}^t \sim \pazocal{N}\left( \mathbf{\mu}^t, \mathbf{\sigma}^t \right)$.
This variable is transformed into a mode via fully connected layers, i.e., $\hat{m}^t = F_{cn} (\mathbf{z}^{t})$.
This process resembles a variational auto-encoder \citep{VAE}; however, our model is supervised to generate conditional signals (modes) instead of reconstructing original input signals (configurations).

\subsection{Following High-Quality Modes}\label{sec:mpc}

Based on MTP, we build MTPnav, a cost-based planner (see \figref{fig:overview}). At planning time $t$, MTPnav invokes MTP, which returns a set of $M$ high-likelihood trajectory predictions $\mathbf{\pazocal{X}}$, corresponding to high-likelihood modes $\vm$ of multiagent behavior. These predictions are then evaluated with respect to a trajectory cost $J:\pazocal{X}^{n}\to\mathbb{R}$:
\begin{equation}
\label{eq:costfn}
    J(\mathbf{X}) = w_{sm}J_{sm}(\mathbf{X}) + w_{ref}J_{ref}(\mathbf{X}) + w_{col}J_{col}(\mathbf{X}) + 
    w_{p}J_{p}(\mathbf{X})\mbox{,}
\end{equation}
where $J_{sm}$ penalizes non-smooth trajectories, $J_{ref}$ penalizes deviations from the ego-vehicle reference trajectory, $J_{col}$ penalizes collisions, $J_{p}$ penalizes low-likelihood trajectories, and $w_{sm}$, $w_{ref}$, $w_{col}$, $w_{p}$ are corresponding weights (refer to Appendix, Sec.~\ref{sec:cost_defn} for detailed formulation). The trajectory of minimum cost $\mathbf{X}_{*}$ is selected and passed to a tracking controller, which returns a control input $u^{*}$ for the agent to execute.

\begin{table}
\centering
\addvbuffer[0pt 2pt]{
\resizebox{\textwidth}{!}{
\begin{tabular}{@{}lcccccc@{}}
\toprule
              Scenario & \multicolumn{2}{c}{Two Agents} & \multicolumn{2}{c}{Three Agents} & \multicolumn{2}{c}{Four Agents} \\ 
              \midrule
              Metric  & minADE (m)$\downarrow$         & minFDE (m)$\downarrow$         & minADE (m)$\downarrow$          & minFDE (m)$\downarrow$         & minADE (m)$\downarrow$         & minFDE (m)$\downarrow$         \\ \midrule
\textbf{MTP}    & \textbf{0.301 $\pm$ 0.346} & \textbf{0.611 $\pm$ 0.893} & \textbf{0.304 $\pm$ 0.459}  & \textbf{0.717 $\pm$ 1.366}     & \textbf{0.264 $\pm$ 0.410}            & \textbf{0.588 $\pm$ 1.067}            \\
MTP-fc          & 0.523 $\pm$ 0.440            & 1.063 $\pm$ 1.020  & 0.456 $\pm$ 0.612 & 1.024 $\pm$ 1.743            & 0.334 $\pm$ 0.342            & 0.704 $\pm$ 0.996 \\
GRU             & 0.929 $\pm$ 0.566         & 1.578 $\pm$ 1.159 & 1.176 $\pm$ 0.727    & 2.119 $\pm$ 1.870            & 0.958 $\pm$ 0.551            & 1.760 $\pm$ 1.408     \\
MFP             & 0.834 $\pm$ 0.969  & 0.825 $\pm$ 1.318  & 1.603 $\pm$ 0.981  & 1.504 $\pm$ 1.857  & 1.556 $\pm$ 0.707 & 1.444 $\pm$ 1.758            \\ \bottomrule
\end{tabular}
}
}
\caption{\textbf{Prediction accuracy} (mean, standard deviation) measured as minADE and minFDE (average and final displacement error) for models trained on datasets with two, three, and four agents. MTP performs significantly better than other models (Wilcoxon signed-rank test, $p<0.001$).
\label{table:pred-eval}}
\end{table}

\section{Evaluation}
\label{sec:evaluation}

We evaluate the performance of the proposed framework in two ways. First, we verify the ability of MTP to produce high-quality, mode-conditioned trajectory reconstruction. Next, we illustrate the value of MTPnav for tasks involving navigation at uncontrolled intersections through a simulated case study on a series of challenging intersection-crossing scenarios.

\subsection{Datasets}

Existing open datasets \citep{Argoverse,lyft2020} do not contain organized and sufficient amounts of interactions at uncontrolled intersections. For this reason, we generated a series of simulated datasets of challenging intersection crossings. Our data-generation pipeline consists of: (1) extracting realistic path references $\tau_{i}$, $i\in \pazocal{N}$ following non-holonomic car kinematics using CARLA \citep{Dosovitskiy17}, and (2) executing these references under a variety of different behaviors and scenarios using a custom, computationally efficient, purely kinematic simulator. Our simulator controls vehicles in a centralized fashion using a priority system that provides  acceleration/deceleration signals based on a time-to-collision heuristic and according to a set of parameters representing agents' behavioral models (desired speed, acceleration, deceleration, etc.). The vehicles track their paths using PID-based steering and speed controllers. This pipeline lets us combine the realism of CARLA with the ability to generate data efficiently and control scenarios and agent behaviors. This strategy also provides the ability to deploy our models directly to CARLA without re-training or fine-tuning to evaluate navigation performance (Sec.~\ref{sec:nav-performance}). Overall, we generate three diverse datasets of the form $\pazocal{D}_{n} =\{\mathbf{X}^{n}_{1}, \dots, \mathbf{X}^{n}_{|\pazocal{D}_{n}|}\}$, containing $|\pazocal{D}_2| = 116k$, $|\pazocal{D}_3| = 1,180M$, and $|\pazocal{D}_4| = 466k$ multiagent trajectories involving $n=2,3,4$ agents, respectively, by methodically varying: (1) agents' destinations $D$ (spanning the set of combinations), (b) their target speeds ($3-12$m/s), and (c) their acceleration/deceleration 
($1-5$m/s$^2$). 

\subsection{Model Training}

We train different trajectory-reconstruction and mode-prediction models on each of the datasets $\pazocal{D}_{2}$, $\pazocal{D}_{3}$, $\pazocal{D}_{4}$, following 9:1 train/test splits.
We partition the datasets into a set of subtrajectories with a length of $H = h_p + h_f$ where $h_p = 15$, and $h_f = 25$.
We transform each subtrajectory into a sequence of graphs $\vg = \left\{ g^{1}, \cdots, g^{H} \right\}$ and split it into $\vg^{p} = \left\{ g^{1}, \cdots, g^{h_{p}} \right\}$ and $\vg^{f} = \left\{ g^{h_{p} + 1}, \cdots, g^{H} \right\}$.
We also compute a corresponding mode signal $m$ for each sequence $\vg$.
Then we train the trajectory reconstruction model with student-forcing, providing the ground-truth mode signal $m$ and a sequence of history graphs $\vg^{p}$ to produce $\hat{\vg}^{f}$.
The mode prediction model is trained with the full ground-truth sequence $\vg$ and generates a mode prediction $\hat{m}$.
\figref{fig:training} depicts an overview of the training process.
We use a mean squared error (MSE) loss for training the trajectory reconstruction model.
For training the mode prediction model, we use a loss similar to the one used by $\beta$-VAE~\citep{Higgins2017betaVAELB} (setting $\beta = 8$), featuring a cross-entropy component for reconstructing the mode signal given the latent variable and a KL-divergence component keeping the latent distribution close to a Gaussian.
Note that at inference time, we only take the predicted mode signal $\hat{m}^{h_p}$ from the mode-prediction model and feed it to the trajectory-reconstruction model to collect $\hat{\vg}^{f}$.
See Sec.~\ref{ssec:appendix-training} for mode details on the training process.

\subsection{Trajectory Prediction Accuracy}

We evaluate the accuracy of MTP in making multimodal trajectory predictions on a set of 1,000 randomly selected subtrajectories from the test sets.
For each subtrajectory, we sample 100 latent variables $\mathbf{z}$ from the predicted distribution and predict trajectories for distinct modes.
We perform an ablation study comparing the performance of MTP to two alternative models: a GRU and a modified MTP without a GRU (MTP-fc).
Both use the same structure to model the latent variable distribution. We also compare MTP to MFP~\citep{tang2019multiple}, which constitutes a state-of-the-art multimodal trajectory prediction framework. We employ the original MFP implementation (fixing the number of generated futures to $k=3$) and modify the code to transform each trajectory to its local coordinate frame, as described by \citet{tang2019multiple} (the state embedding in their implementation differs from the original paper). Note that other models (including MTP) do not use a local transformation.

Table \ref{table:pred-eval} shows the prediction accuracy of our framework, measured using the minADE and minFDE metrics (the minimum values of average/final displacement errors in multiple predictions).
Our model outperforms the baselines across all datasets.
It exhibits on average 71.63\% lower minADE error than GRU and 33.82\% less than MTP-fc.
Importantly, our model outperforms MFP by 78.24\% (on average) in terms of minADE.
Further, note that MTP maintains comparable errors as the number of agents increases; in contrast, the performance of MFP drops.
\textit{Our insight is that the superior performance achieved by MTP can be attributed to the salient encoding of interaction provided by the formal definition of \emph{modes}.}
Our model can leverage physically meaningful signals as conditional signals instead of choosing one of a fixed number of latent behaviors.
However, note that MFP was designed to apply more broadly to a variety of driving environments, unlike our approach, which was specifically designed for intersections.

\subsection{Navigation Performance}\label{sec:nav-performance}

\begin{figure}
\centering
\begin{subfigure}{0.24\linewidth}
\centering
\includegraphics[width=\linewidth]{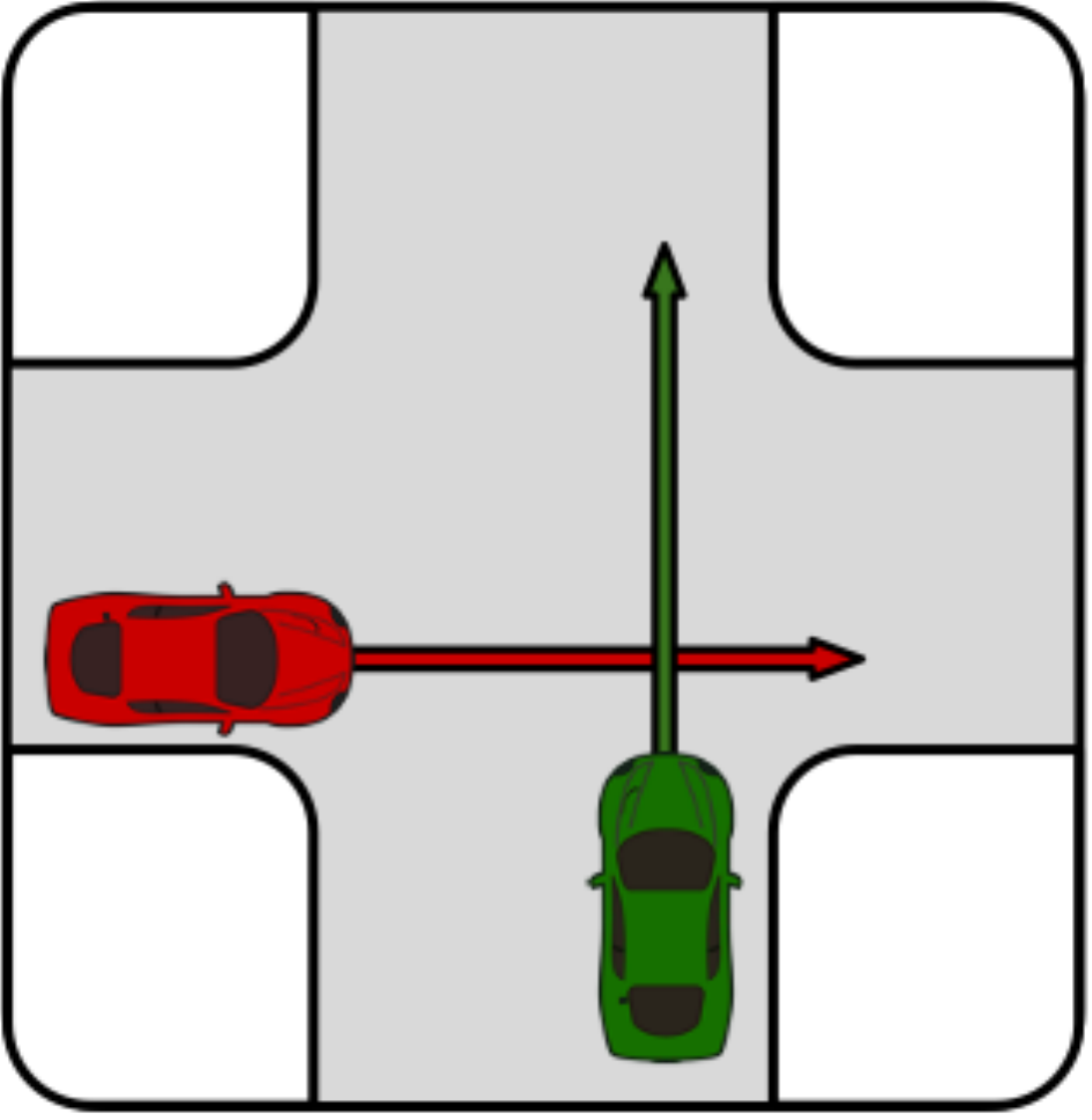}
\caption{Two-agent scenario.\label{fig:scenario2agents}}
\end{subfigure}
~
\begin{subfigure}{0.24\linewidth}
\centering
\includegraphics[width=\linewidth]{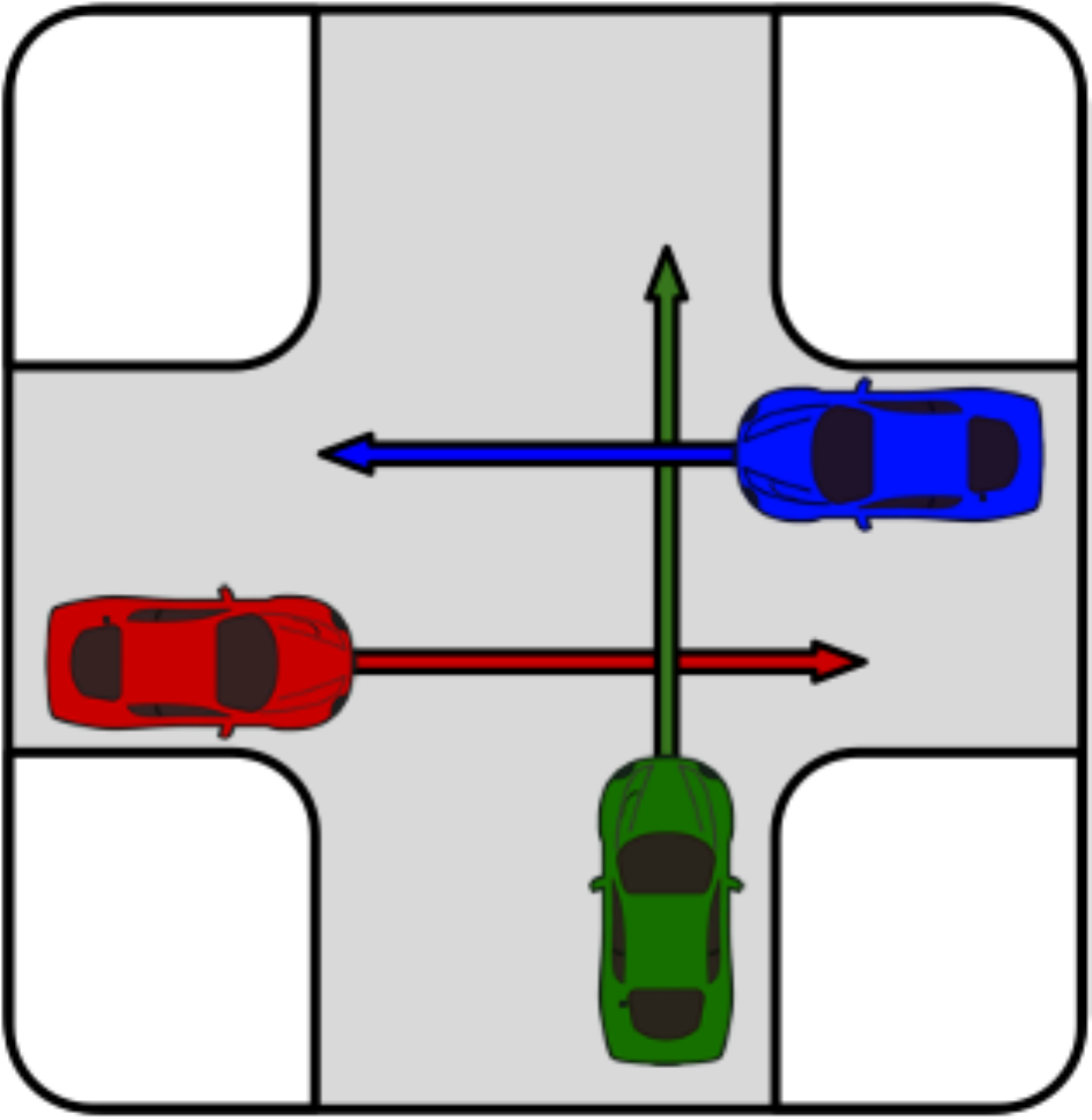}
\caption{Three-agent scenario.\label{fig:scenario3agents}}
\end{subfigure}
~
\begin{subfigure}{0.24\linewidth}
\centering
\includegraphics[width=\linewidth]{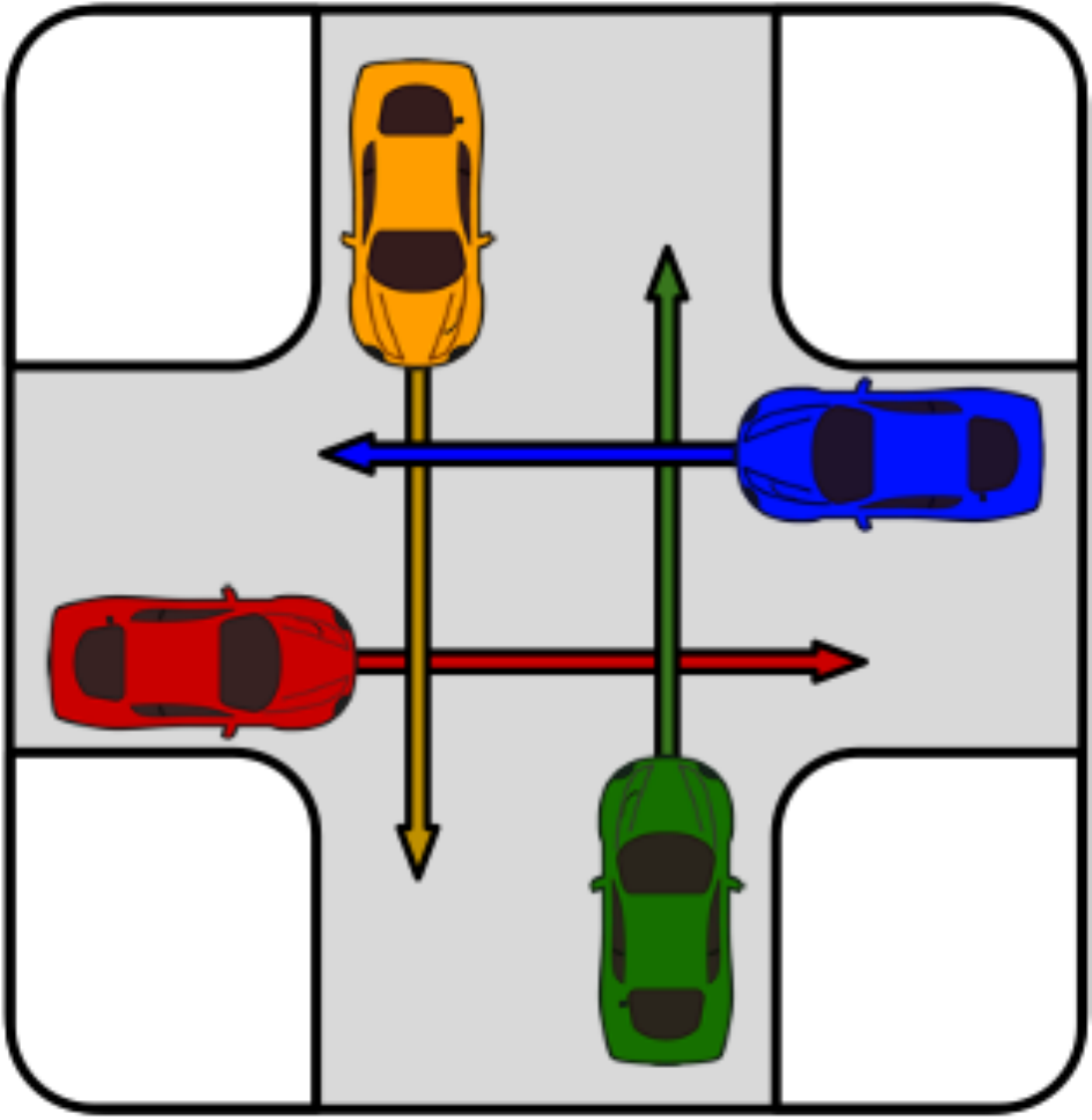}
\caption{Four-agent scenario.\label{fig:scenario4agents}}
\end{subfigure}
\caption{Navigation scenarios considered in our evaluation on CARLA.\label{fig:eval_scenarios}}
\end{figure}

We further evaluate the quality of the MTP predictions through a simulated case study on a decentralized navigation task at a four-way uncontrolled intersection of size $60 \times 60$~\SI{}{\meter} on CARLA \citep{Dosovitskiy17}. We consider three different intersection-crossing scenarios involving $n=2,3,4$ agents, specifically selected to elicit challenging vehicle interactions, as shown in \figref{fig:eval_scenarios}. During execution, the ego agent (agent $\#1$, the bottom agent in \figref{fig:eval_scenarios}) is running our framework (MTPnav), while other agents are running the CARLA \citep{Dosovitskiy17} Autopilot controller.\footnote[1]{Autopilot is a widely available baseline that is  part of the CARLA simulator \citep{Dosovitskiy17}.} We consider two different conditions: \emph{Easy} and \emph{Hard}. \emph{Easy} consists of \emph{Cautious} and/or \emph{Aggressive} agents with target speeds sampled uniformly from $[10,25]$ and $[30,45]$ ($km/h$), respectively. \emph{Hard} consists of \emph{Aggressive} and/or \emph{Normal} agents with target speeds sampled from $[30,45]$ and $[20,40]$ ($km/h$), respectively, where \emph{Cautious}, \emph{Normal} and \emph{Aggressive} are behavioral agents provided by CARLA. Uniformly sampling from these ranges, we construct 50 random experiments per scenario and condition. 

We compare the performance of MTPnav to two baselines: the Autopilot (homogeneous case, serving as a reference for the difficulty of the task) and a purely reactive Model Predictive Controller (MPC) that treats other agents as obstacles. We employ using two metrics: (1) collision frequency $\pazocal{C} (\%)$ and (2) the time the ego-agent took to reach its destination $\pazocal{T}$ ($s$). If the ego-agent collides, we mark the trial as \emph{in-collision} and assign it a time penalty of $30s$ corresponding to the time-to-destination for a \emph{Cautious} agent with constant speed of $10km/h$. Trials that did not terminate by the $30s$ mark are also marked as in-collision.


Table~\ref{tab:results} shows the performance of MTPnav. We observe that MTPnav outperforms the baselines across all scenarios and interaction settings in terms of both time and collision frequency. In particular, its collision frequency is close to zero for scenarios with two and three agents, and significantly lower in scenarios involving four agents. We also see that it is significantly more time efficient, especially in the \emph{Hard} scenarios. Overall, we attribute MTPnav's performance to its ability to exploit the domain structure: MTPnav agents are capable of rapidly adapting to the dynamic environment by foreseeing the multiagent interaction dynamics. We see a significant increase in collisions for scenarios involving four agents. This could be attributed to the steep increase in domain complexity (for reference, the number of possible modes is $162$) but also possibly to the relatively small four-agent dataset that we employed.


\begin{table}
\centering
\resizebox{\columnwidth}{!}{%
\addvbuffer[0pt 2pt]{
\begin{tabular}{lccccccc}
\toprule
\multicolumn{1}{l}{Scenario} & & \multicolumn{2}{c}{Two Agents} & \multicolumn{2}{c}{Three Agents} & \multicolumn{2}{c}{Four Agents} \\ 
\midrule
Interaction & & Easy & Hard & Easy & Hard & Easy & Hard\\


\midrule
\multicolumn{1}{l|}{\multirow{2}{*}{Autopilot}} & \multicolumn{1}{c|}{$\pazocal{T} (s)$} & \multicolumn{1}{c|}{$\textcolor{red}{23.37 \pm 6.01}$} &
\multicolumn{1}{c|}{$\textcolor{red}{26.00 \pm 7.52}$} &
\multicolumn{1}{c|}{$23.01 \pm 5.78$} &
\multicolumn{1}{c|}{$\textcolor{red}{28.69 \pm 3.96}$} &
\multicolumn{1}{c|}{$\textcolor{teal}{22.16 \pm 7.56}$} &
\multicolumn{1}{c}{$\textcolor{red}{28.41 \pm 4.31}$}\\

\multicolumn{1}{c|}{} & \multicolumn{1}{c|}{$\pazocal{C} (\%)$} & \multicolumn{1}{c|}{$\textcolor{red}{28.00\pm 6.35}$} &
\multicolumn{1}{c|}{$\textcolor{red}{78.00\pm 5.86}$} &
\multicolumn{1}{c|}{$\textcolor{red}{30.00\pm 6.48}$} &
\multicolumn{1}{c|}{$\textcolor{red}{86.00\pm 4.91}$} &
\multicolumn{1}{c|}{$\textcolor{red}{40.00\pm 6.93}$} &
\multicolumn{1}{c}{$\textcolor{red}{88.00\pm 4.60}$}\\ \midrule

\multicolumn{1}{l|}{\multirow{2}{*}{MPC}} & \multicolumn{1}{c|}{$\pazocal{T} (s)$} & \multicolumn{1}{c|}{$\textcolor{blue}{22.37 \pm 6.26}$} &
\multicolumn{1}{c|}{$\textcolor{red}{26.98 \pm 6.43}$} &
\multicolumn{1}{c|}{$21.34 \pm 6.26$} &
\multicolumn{1}{c|}{$\textcolor{red}{27.78 \pm 5.54}$} &
\multicolumn{1}{c|}{$\textcolor{teal}{21.28 \pm 7.98}$} &
\multicolumn{1}{c}{$\textcolor{red}{28.25 \pm 4.74}$}\\

\multicolumn{1}{c|}{} & \multicolumn{1}{c|}{$\pazocal{C} (\%)$} & \multicolumn{1}{c|}{$\textcolor{red}{30.00\pm 6.48}$} &
\multicolumn{1}{c|}{$\textcolor{red}{82.0\pm 5.43}$} &
\multicolumn{1}{c|}{$\textcolor{red}{28.00\pm 6.35}$} &
\multicolumn{1}{c|}{$\textcolor{red}{86.00\pm 4.91}$} &
\multicolumn{1}{c|}{$\textcolor{red}{40.0\pm 6.93}$} &
\multicolumn{1}{c}{$\textcolor{red}{88.0\pm 4.60}$}\\ \midrule

\multicolumn{1}{l|}{\multirow{2}{*}{\textbf{MTPnav}}} & \multicolumn{1}{c|}{$\pazocal{T} (s)$} & \multicolumn{1}{c|}{$\mathbf{19.13 \pm 3.78}$} &
\multicolumn{1}{c|}{$\mathbf{14.81 \pm 1.56}$}&
\multicolumn{1}{c|}{$\mathbf{20.85 \pm 3.71}$} &
\multicolumn{1}{c|}{$\mathbf{18.70 \pm 0.61}$} &
\multicolumn{1}{c|}{$\mathbf{19.34 \pm 8.03}$} &
\multicolumn{1}{c}{$\mathbf{21.36 \pm 5.91}$}\\

\multicolumn{1}{c|}{} & \multicolumn{1}{c|}{$\pazocal{C} (\%)$} & \multicolumn{1}{c|}{$\mathbf{2.00\pm 1.98}$} &
\multicolumn{1}{c|}{$\mathbf{0}$} &
\multicolumn{1}{c|}{$\mathbf{2.00\pm 1.98}$} &
\multicolumn{1}{c|}{$\mathbf{0}$} &
\multicolumn{1}{c|}{$\mathbf{28.00\pm 6.35}$} &
\multicolumn{1}{c}{$\mathbf{30.00\pm 6.48}$} \\

\bottomrule
\end{tabular}
}
}
\caption{\textbf{Navigation performance measured with respect to collision frequency ($\pazocal{C}$) and time to destination for ego-agent} ($\pazocal{T}$). Each entry contains a mean and a standard deviation over 50 trials. red, green and blue entries indicate scenarios in which MTPnav outperformed other methods at significance levels $p<0.001$, $p<0.01$, $p<0.05$ respectively (Wilcoxon signed-rank test).}
\label{tab:results}
\end{table}

\section{Discussion}\label{sec:discussion}

We introduced a prediction model (MTP) and a planner (MTPnav) for navigation at uncontrolled intersections, leveraging a mathematical insight about the topological structure of intersection crossings. MTP, trained on datasets acquired using a custom-built, lightweight simulator, outperformed a state-of-the-art,  trajectory-prediction baseline \citep{tang2019multiple} and enabled MTPnav to exhibit safe and time-efficient behavior across a series of challenging intersection-crossing tasks (without additional fine-tuning or retraining) that were conducted using the high-fidelity simulation engine CARLA. Our findings illustrate the value of reasoning about topological modes as a strategy to guide prediction towards a small representative set of outcomes. Ongoing work involves real-world experiments on a miniature racecar \citep{srinivasa2019mushr} to highlight the virtues of our approach for real-world applications. 

Our work is not without limitations. Our data generation and evaluation pipelines are constrained by our behavior parametrizations and the mechanics of our simulators. Further, our evaluation experiments are limited by the computational load of running expensive CARLA experiments. Future work will focus on leveraging our empirical insights towards expanding our simulator to capture richer spaces of behaviors for data generation and testing. Finally, the MTPnav framework is proposed as an example, non-optimized incorporation of MTP on a cost-based planner; alternative use cases could apply reinforcement learning or model predictive control frameworks.



\acknowledgments{This work was (partially) funded by the National Science Foundation IIS (\#2007011), National Science Foundation DMS (\#1839371), the Office of Naval Research, US Army Research Laboratory CCDC, Amazon, Honda Research Institute USA, UW Reality Lab, Facebook, Google, Huawei, and a Samsung Scholarship.
We would also like to thank NVIDIA for generously providing DGX via the NVIDIA Robotics
Lab and the UW NVIDIA AI Lab (NVAIL).
}

\bibliography{references}
\appendix
\section*{Appendix}
\renewcommand{\thesubsection}{\Alph{subsection}}
\setcounter{table}{0}
\renewcommand{\thetable}{A\arabic{table}}

\label{sec:appendix}
\subsection{Cost functions definition}\label{sec:cost_defn}
We now provide the formulation for individual cost functions used in Eq.~\ref{eq:costfn}:
\begin{center}
\bgroup
\def\arraystretch{1.8}
\begin{tabular}{l}
    \text{Smoothness cost}: $J_{sm}(\mathbf{X}) = \sum\limits_{t=0}^{T - 1}\gamma^{t}||v_{t+1} - v_{t}||$\mbox{,}\\
    \text{Cross-track error}: $J_{ref}(\mathbf{X}) = dist(\tau^{t}, X^{t}) * sin(\psi_{\tau^{t}} - \psi_{X^{t}})$\mbox{,} \\
    Collision cost: $J_{col}(\mathbf{X}) = \sum\limits_{t=0}^{T}\gamma^{t}c_{t}(X^{t})$, where
        \begin{math}
        c_{t}(X^{t}) = 
        \begin{cases}
        0 & \textnormal{if}\: D_{i}(X^{t}) \geq d_{min}\\
        1 & \textnormal{else}
        \end{cases}\mbox{,}
        \end{math}\\
    Likelihood cost: $J_{p}(\mathbf{X}) = \frac{\displaystyle 1}{\displaystyle p_{m_i}}$ \\
\end{tabular}
\egroup
\end{center}

where $v_{t}$ is the velocity at timestep \textit{t}. $dist(\tau^{t}, X^{t})$ is the euclidean distance between agent position and reference waypoint at timestep \textit{t}. $\psi_{\tau^{t}}$ is the heading angle on reference waypoint and $\psi_{X^{t}}$ is the heading angle of the agent, at timestep \textit{t}. $p_{m_{i}}$ is the likelihood for mode $m_{i}$. $D_{i}$ is a distance function and $d_{min}$ is the distance threshold for collision. To reduce the computational burden of collision-checking, we use the circle-based collision checking method by reducing the car footprint to a set of three circles covering the car volume. Any object is thus in collision with the vehicle, if its distance from the center is less than the specified threshold. 

\subsection{Implementation Details}\label{sec:implementation}

In this section, we provide more details about the data generation pipeline, the training process and the navigation experiments.

\subsubsection{Data generation}

We generated three different datasets using our custom simulator by varying three simulation parameters: a) agents' configurations (combinations of starting locations and intended destinations); b) agents' target speed; c) agents' acceleration/deceleration.

\begin{itemize}
    \item Two agents
    \begin{itemize}
        \item Number of configurations: $N_{d2} = 3^3=27$ (taking all possible configurations)
        \item Number of speeds: $N_{s2} = 7^2$ (sampled from $[2.8-12.5]$ m/s with step size 1.4 m/s)
        \item Number of accelerations: $N_{a2} = 10^2$ (sampled uniformly from $[1-5]$) m/s
        \item Number of total episodes: $N_{e2} = N_{d2}\times N_{s2}\times N_{a2} \approx 132K$
    \end{itemize}
    \item Three agents
    \begin{itemize}
        \item Number of configurations: $N_{d3} = N_{d2}\times 2\times 3 = 162$
        \item Number of speeds: $N_{s3} = 7^3$ (sampled from $[2.8-12.5]$ m/s, step size 2.8 m/s)
        \item Number of accelerations: $N_{a3} = 5^3$ (sampled uniformly from $[1-5]$) m/s
        \item Number of total episodes: $N_{e3} = N_{d3}\times N_{s3}\times N_{a3} \approx 1,29M$
        
    \end{itemize}
    \item Four agents
    \begin{itemize}
        \item Number of configurations: $N_{d4} = 3^4$
        \item Number of speeds: $N_{s4} = 3^4$ (sampled from $[2.8-9.7]$ m/s, step size 2.8 m/s)
        \item Number of accelerations: $N_{a4} = 3^4$ (sampled uniformly from $[1-5]$) m/s
        \item Number of total episodes: $N_{e4} = N_{d4}\times N_{s4}\times N_{a4} \approx 531K$
    \end{itemize}
\end{itemize}

The final datasets were extracted upon pruning out episodes including collisions (see main paper for dataset sizes).

\subsubsection{Training}
\label{ssec:appendix-training}
\textbf{Dataset split}
We split the graph sequence $\vg = g^{1:H}$ into a sequence of history graphs $\vg^p = g^{1:h_p}$ and a sequence of target graphs $\vg^f = g^{h_p+1:H}$.

\textbf{Trajectory-reconstruction model training}
The trajectory-reconstruction model takes a graph $g^t$ and a hidden graph $g^t_h$ and predicts a graph at next timestep $\hat{g}^{t+1}$ along with an updated hidden graph $g_h^{t+1}$ (see \figref{fig:traj-recon-model}).
Training procedure has two stages: reading history sequence and predicting future sequence.
In the first stage where $t < h_p$, the model takes $g^t$ from the history graphs $\vg^p$ and the output graph in this stage is discarded.
Note that the hidden graph remains updated to keep the information propagated through time.
In the second stage where $t \geq h_p$, the input graph is taken from the output graph at the previous timestep $t-1$, $\hat{g}^{t}$.
The output graphs are collected to form the future trajectory.
This process is repeated until the full predicted graph sequence $\hat{\vg}^{f} = \hat{g}^{h_{p}+1:H}$ is produced.
Loss is computed on $\hat{\vg}^f$ and $\vg^f$ by a mean squared error function.
In the training, we provide ground-truth mode signals.

\textbf{Mode-prediction model training}
The mode-prediction model shares a similar architecture (see \figref{fig:mode-pred-model}), with the difference that we provide the ground-truth sequence of graphs $\vg$ to the model and collect mode signals $\hat{m}^{t}$ for $t \geq h_p$.

\textbf{Hyperparameters}
We trained the models considered (both MTP and MFP) using PyTorch. For \textbf{MTP}, we set the following options -- (Optimizer: Adam; Learning rate: $1 \times 10^{-3}$; Gradient clip: 1.0; Hidden unit size: 30. For \textbf{MFP}, we made use of the official implementation (\url{https://github.com/apple/ml-multiple-futures-prediction}) setting the following options -- (Number of modes: 3; Subsampling: 2; Encoder size: 64; Decoder size: 128; Neighbor encoding size: 8; Neighbor attention embedding size: 20; Remove $y$ mean: False; Use forcing: classmate forcing).

\subsection{Experiment Setup in CARLA}
\begin{figure}
    \centering
    \includegraphics[width=0.9\linewidth]{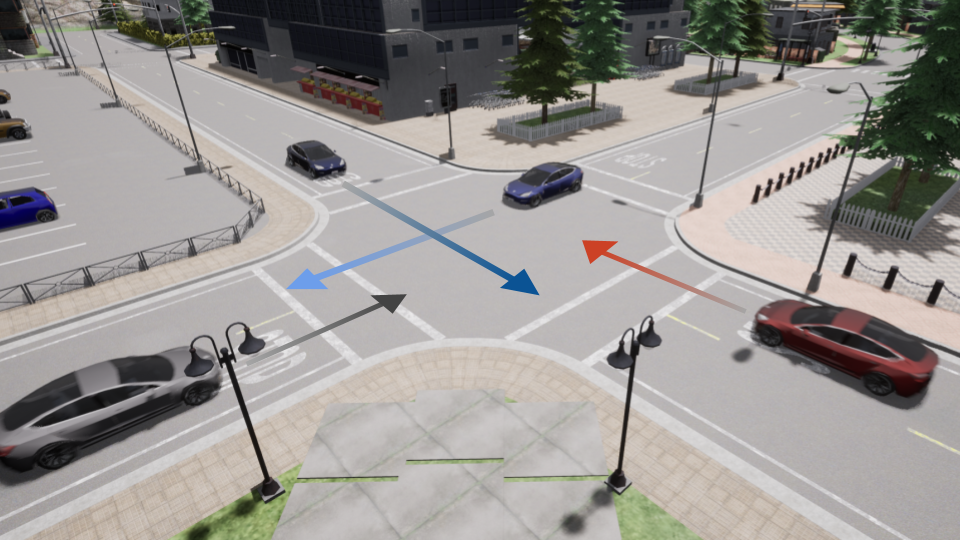}
    \caption{An example of the 4-agent scenario at an uncontrolled intersection in CARLA}
    \label{fig:carla_preview}
\end{figure}

The navigation experiments are conducted at an uncontrolledf intersection within the \texttt{Town04} map of CARLA. For each experiment, agents are spawned on a specified side of the intersection and headed towards a destination waypoint corresponding to an intended direction (left, right, forward). At the beginning, all the agents are controlled using the Autopilot provided by CARLA. The ego agent controller switches to our method (MTPnav) or one of the baselines when the ego-agent reaches a distance of 25m from the center of the intersection, and switches back to Autopilot after crossing the intersection. The weights used in the cost function (see main paper) are set to the following values: $w_{sm} = 50$, $w_{ref} = 100$, $w_{col} = 10000$, $w_{p} = 1$.

\end{document}